\documentclass[letterpaper]{article} 
\usepackage{aaai2026}  
\usepackage{times}  
\usepackage{helvet}  
\usepackage{courier}  
\usepackage[hyphens]{url}  
\usepackage{graphicx} 
\usepackage{natbib}  
\usepackage{caption} 
\usepackage{natbib}
\usepackage{booktabs}
\usepackage[table]{xcolor}
\usepackage{amssymb}
\usepackage{multirow}
\usepackage{amssymb}
\usepackage{amsmath}
\usepackage{algorithm}
\usepackage{algorithmic}
\usepackage{adjustbox}
\usepackage{xcolor}
\usepackage{newfloat}
\usepackage{listings}
\DeclareCaptionStyle{ruled}{labelfont=normalfont,labelsep=colon,strut=off} 
\floatstyle{ruled}
\newfloat{listing}{tb}{lst}{}
\floatname{listing}{Listing}
\title{Learnable Permutation for Structured Sparsity on Transformer Models}
\author{
    Zekai Li\equalcontrib, Ji Liu\equalcontrib, Guanchen Li, Yixing Xu, Ziqiong Liu, Xuanwu Yin, Dong Li, Emad Barsoum
}
\affiliations{
    Advanced Micro Devices, Inc.\\
    \{Zekai.Li, Ji.Liu, Guanchen.Li, Yixing.Xu, Ziqiong.Liu, Xuanwu.Yin, d.li, Emad.Barsoum\}@amd.com
}

\usepackage{bibentry}

\begin{document}

\maketitle

\begin{abstract}
Structured sparsity has emerged as a popular model pruning technique, widely adopted in various architectures, including CNNs, Transformer models, and especially large language models (LLMs) in recent years. A promising direction to further improve post-pruning performance is weight permutation, which reorders model weights into patterns more amenable to pruning. However, the exponential growth of the permutation search space with the scale of Transformer architectures forces most methods to rely on greedy or heuristic algorithms, limiting the effectiveness of reordering.

In this work, we propose a novel \textbf{end-to-end learnable} permutation framework. Our method introduces a learnable permutation cost matrix to quantify the cost of swapping any two input channels
of a given weight matrix, a differentiable bipartite matching solver to obtain the optimal binary permutation matrix given a cost matrix, and a sparsity optimization loss function to directly optimize the permutation operator.
We extensively validate our approach on vision and language Transformers, demonstrating that our method achieves state-of-the-art permutation results for structured sparsity.
\end{abstract}


\section{Introduction}
Transformer architectures~\cite{c:22} have achieved remarkable success across diverse AI applications, including vision models such as ViT~\cite{yuan2021tokens}, DETR~\cite{carion2020end}, DiT~\cite{peebles2023scalable}, and large language models (LLMs) such as GPT~\cite{floridi2020gpt}, LLaMA~\cite{touvron2023llama}, Qwen~\cite{bai2023qwen}, and DeepSeek~\cite{guo2025deepseek}. Their strong representational capacity and generalizability have made Transformers the preferred architecture for foundation models. However, deploying these large-scale models on resource-constrained hardware remains challenging, as inference cost grows rapidly with increasing model size. To this end, structured pruning under N:M sparsity constraints, which requires that only N out of every group of M consecutive weights remain nonzero, has emerged as an efficient solution~\cite{bengio2013estimating, han2015learning, sun2023wanda, fang2024maskllm}. Its regular structure enables significant parameter reduction while maintaining hardware compatibility, as demonstrated by recent GPUs that accelerate structured sparse patterns such as 2:4 sparsity~\cite{zhou2021learning}.

Despite their effectiveness, structured pruning methods such as Wanda~\cite{sun2023simple}, SparseGPT~\cite{frantar2023sparsegpt}, and PrunerZero~\cite{dong2024pruner} still degrade accuracy due to a fundamental mismatch between rigid sparsity patterns and inherent weight distributions. Standard N:M pruning preserves only the top-N weights within fixed-size groups, irrespective of actual weight importance. Since Transformer channels are initially ordered arbitrarily, important weights can easily be pruned unintentionally. To mitigate this, channel-wise weight permutation methods reorder weight matrices before pruning to better align weight saliency with sparsity patterns, significantly reducing accuracy loss~\cite{pool2021channel}.

However, current permutation approaches mainly rely on heuristic or greedy algorithms~\cite{zhangplug}, which optimize local importance scores rather than directly improving end-to-end task performance. Furthermore, these heuristics neglect global coordination and are computationally expensive, typically employing costly linear sum assignment or searching algorithms~\cite{pool2021channel}. Such complexity becomes impractical for large Transformer models with numerous layers and channels, limiting the efficiency and quality of resulting permutations.

To address these shortcomings, we propose a fully learnable permutation framework to jointly optimize channel permutation and structured pruning in an end-to-end manner. However, there are two significant challenges. \textbf{First, the permutation operation is inherently discrete and non-differentiable}, complicating its integration with gradient-based training. \textbf{Second, existing importance heuristics are insufficient for guiding permutation decisions that affect overall model performance.} This calls for an end-to-end optimization framework that directly links permutation learning with task-level objectives.
In response, our framework introduces three key innovations:

\begin{itemize}
    \item A \textbf{learnable permutation cost matrix} that explicitly quantifies the cost of swapping any two input channels of a given weight matrix.
    \item To address the non-differentiability of discrete permutation, we design a \textbf{differentiable approximation of bipartite matching} guided by the learnable cost matrix, enabling efficient and accurate binary permutation matrix learning with minimal computational overhead.
    \item An \textbf{end-to-end sparsity optimization loss} function is proposed to jointly guide the optimization of the permutation operator, achieving a fine balance between task-specific performance and alignment with the dense teacher model through knowledge distillation.
\end{itemize}

Through end-to-end learning, the proposed framework derives a dedicated permutation matrix for each weight tensor, which is then multiplied with the original weights to produce reordered weights that align more naturally with the target sparsity pattern.
We apply this approach to both vision and language models, including ViT, LLM, and VLM backbones, and conduct extensive experiments to validate its effectiveness. Experimental results demonstrate that our framework achieves state-of-the-art structured sparsity with significantly reduced accuracy degradation, outperforming traditional greedy baselines on a variety of benchmarks.
\section{Related Work}
\textbf{Model Pruning.}
Model pruning compresses a pre-trained model by reducing its parameter count, memory usage, and computational footprint~\cite{li2025gumiho}. Contemporary pruning approaches can be broadly categorized into three types. 
\textit{Unstructured pruning} eliminates individual weight elements, offering fine-grained sparsity control. However, the resulting irregular sparsity patterns pose challenges for hardware acceleration, often requiring extremely high sparsity levels to achieve meaningful speedup~\cite{han2015deep_compression, han2015learning, liao2025spark}.
\textit{Structural pruning} removes entire filters, channels, or layers, producing regular sparsity patterns that are hardware friendly. While it simplifies deployment, this coarse-grained pruning often leads to considerable accuracy degradation and typically demands retraining to recover performance~\cite{ma2023llmpruner, xia2023sheared, he2017channel}.
\textit{(Semi-)Structured pruning}, or N:M sparsity, enforces a fixed number of nonzero weights per block, balancing accuracy preservation with hardware efficiency. It retains much of the flexibility of unstructured pruning while producing regular memory layouts suitable for modern accelerators~\cite{pool2021accelerating, pool2021channel, frantar2023sparsegpt}.

\textbf{N:M Sparsity.}
The N:M sparsity constraint enforces at most N nonzero values within each block of M weights, achieving a favorable trade-off between compression and inference efficiency on sparsity-aware hardware~\cite{pool2021accelerating, pool2021channel, fang2024maskllm, hu2024s}. 
Earlier methods applied static pruning masks after training, while recent techniques integrate mask learning into the optimization loop using continuous relaxations and gradient-based updates~\cite{zhou2021learning, lu2023step, fang2024maskllm, liuproxsparse}. Approaches such as sparse-refined straight-through estimators further promote the retention of important weights while maintaining strict adherence to the N:M sparsity constraint~\cite{bengio2013estimating, han2015learning}.
Beyond mask selection, post pruning weight update methods recover accuracy under the N:M constraint by solving local reconstruction subproblems via second order updates or constrained quadratic optimization~\cite{frantar2023sparsegpt, bovza2024fast}. 
Our work is orthogonal to these pruning and weight update techniques. We focus on learning channel permutations that align saliency with the N:M mask, which can be integrated seamlessly.

\textbf{Matrix Permutation for Pruning Optimization.}
Matrix permutation aims to rearrange weights such that salient and non-salient values are distributed more uniformly across pruning groups. This improves alignment with structured sparsity patterns like N:M, enhancing pruning compatibility and preserving model performance.
Channel permutation was first introduced in~\cite{pool2021channel}, which identifies an optimal reordering via exhaustive greedy search. However, such methods become impractical when applied to large language models due to the computational cost of processing high-dimensional weight matrices. The Plug-and-Play method~\cite{zhang2024plug} formulates permutation as a combinatorial optimization problem and solves it efficiently using the Hungarian algorithm. However, as a rule-based method, it does not support end-to-end optimization and incurs high cost in large-scale models due to the linear sum assignment operation over large weight tensors.

\vskip 1em
\noindent In this work, we tackle the large search space of channel permutations by introducing a learnable permutation mechanism for GPT-scale Transformers. Our method enables effective channel reordering that improves model accuracy under N:M sparsity constraints.
\section{Methods}

\begin{figure}[t]
\centering
\includegraphics[width=0.46\textwidth]{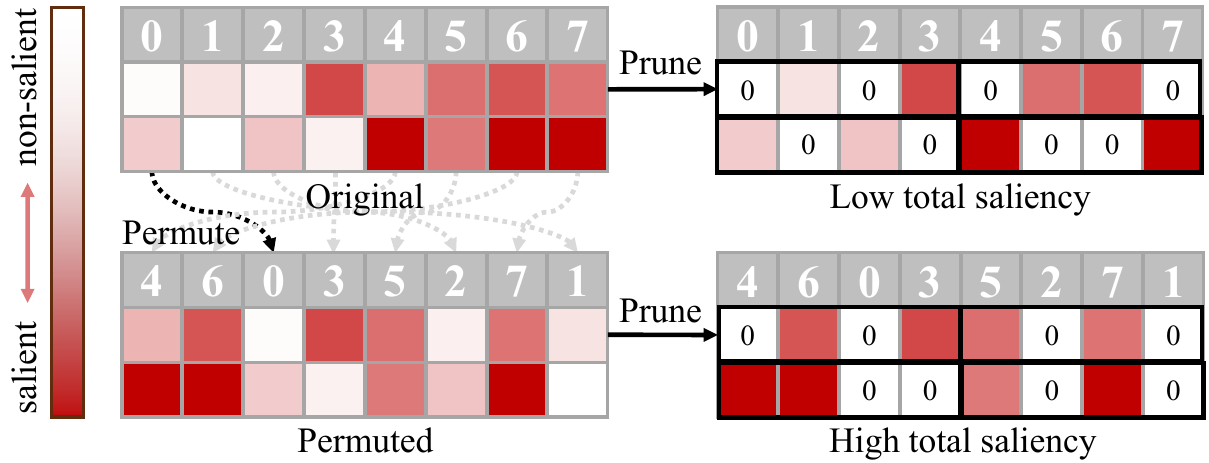} 
\caption{The channel permutation process enhances the friendliness of the 2:4 sparsification, making the overall saliency of the pruning metric more preserved.}
\label{fig:channel-perm}
\end{figure}

We begin by introducing preliminaries on channel permutation in the context of optimizing structured N:M sparsity for Transformers. We then describe our proposed end-to-end learnable permutation framework, which consists of a learnable cost prediction module, a differentiable bipartite matching solver, and  optimization objectives. 
The unified framework facilitates end-to-end optimization of permutation operators for diverse Transformer-based architectures, including vision, language, and vision-language models.

\subsection{Preliminaries}
\label{sec:preliminaries}

\begin{figure}[t]
\centering
\includegraphics[width=0.46\textwidth]{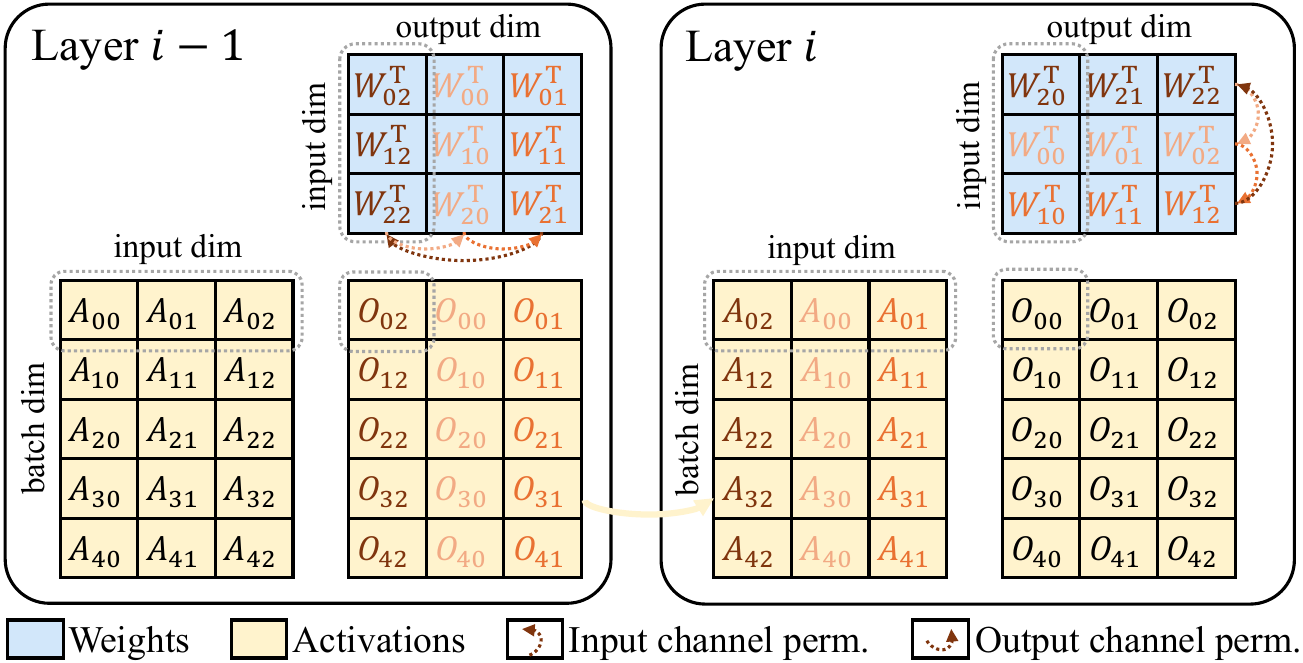} 
\caption{Channel permutation for Linear layers. To guarantee output consistency, after the input channel of the $i$-th layer weight is permuted, the input activation of that layer should also be permuted accordingly, which can be realized by permuting the output channel of the previous ($i-1$)-th layer weight accordingly.}
\label{fig:perm_between_layers}
\end{figure}

\paragraph{Optimize N:M Sparsity via Channel Permutation.} 
Channel permutation enhances the compatibility of weight tensors with structured sparsity patterns such as N:M sparsity. As shown in Figure~\ref{fig:channel-perm}, the weight layout often exhibits uneven saliency, with important weights clustered within certain groups. This skewed distribution lowers the chance of retaining key weights under fixed 2:4 sparsity, resulting in suboptimal pruning with reduced preserved saliency.

By permuting channels before pruning, saliency becomes more evenly distributed across groups. This increases the likelihood that each M-element group contains a mix of important and unimportant weights, allowing structured pruning to retain more informative elements. Channel permutation thus improves N:M pruning by aligning weight layout with sparsity constraints.

\paragraph{Channel Permutation for Linear Layers.} 
Applying a channel permutation to a linear layer's input dimension requires aligning the permuted weights with its input activation to preserve output consistency. Let \(\mathbf{W}_i^{\top} \in \mathbb{R}^{d_\text{in} \times d_\text{out}}\) be the weight matrix of the \(i\)-th linear layer, and let \(\mathbf{P}\in \{0,1\}^{d_\text{in} \times d_\text{in}}\) be a permutation matrix. Applying input channel permutation to the weight yields \(\widehat{\mathbf{W}}_i^{\top} = \mathbf{P} \mathbf{W}_i^{\top}\). To maintain correct computation, the corresponding activation \(\mathbf{A}_i \in \mathbb{R}^{d_\text{batch} \times d_\text{in}}\) must also be transformed as \(\widehat{\mathbf{A}}_i = \mathbf{A}_i \mathbf{P}^{\top}\), so that the output remains unchanged (\(\mathbf{P}^{\top} = \mathbf{P}^{-1}\)):

\begin{equation}
   \widehat{\mathbf{A}}_{i+1} = \widehat{\mathbf{A}}_i \widehat{\mathbf{W}}_i^{\top} = \mathbf{A}_i \mathbf{P}^{\top} \mathbf{P} \mathbf{W}_i^{\top} = \mathbf{A}_i \mathbf{W}_i^{\top} = \mathbf{A}_{i+1}, 
   \label{eq:1}
\end{equation}


To avoid runtime activation permutation, we propagate the permutation backward to the output channels of the preceding \((i{-}1)\)-th layer (\(\widehat{\mathbf{W}}_{i-1} = \mathbf{W}_{i-1} \mathbf{P}^{\top}\)), as shown in Figure~\ref{fig:perm_between_layers}, so that:

\begin{equation}
    \widehat{\mathbf{A}}_i = \mathbf{A}_{i-1} \widehat{\mathbf{W}}_{i-1} = \mathbf{A}_{i-1} \mathbf{W}_{i-1} \mathbf{P}^{\top} = \mathbf{A}_i \mathbf{P}^{\top}.
    \label{eq:2}
\end{equation}

\begin{figure}[t]
\centering
\includegraphics[width=0.46\textwidth]{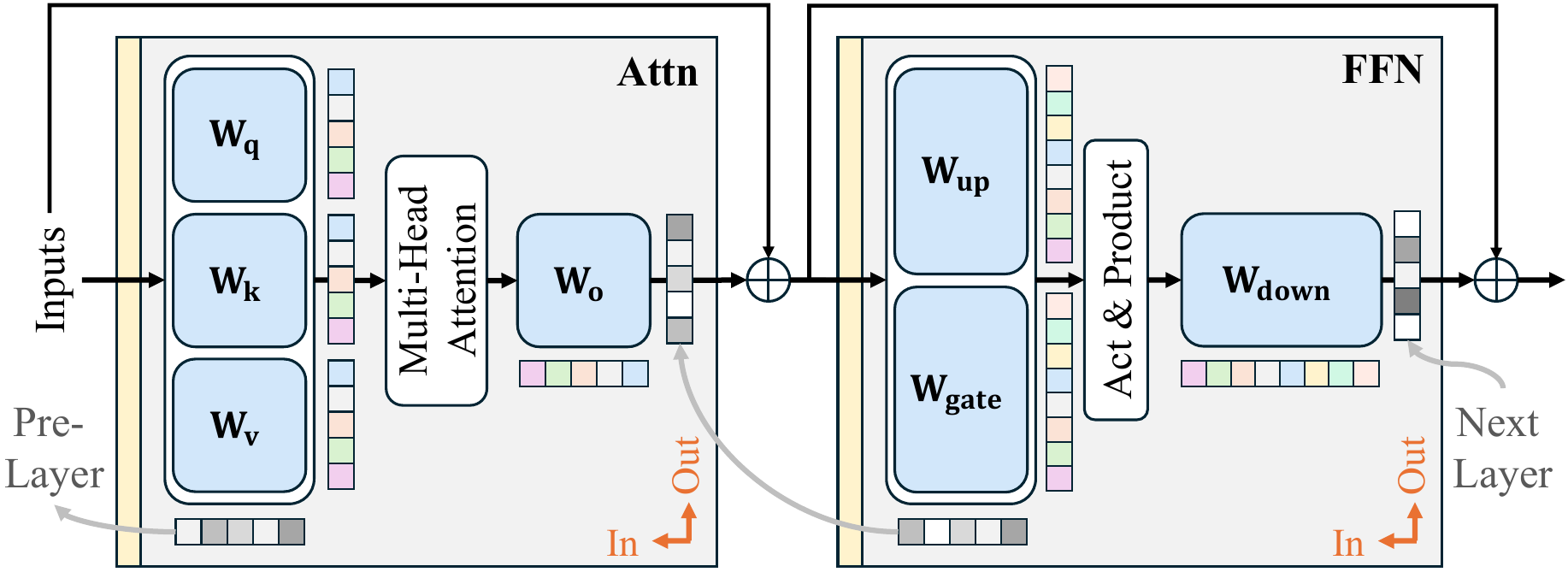} 
\caption{An overview of channel permutation for Transformer layers. The alignment of the input channel permutation of the current layer's weights to the output dimension of the previous layer's weights reflects a structural coupling.}
\label{fig:transformer_perm}
\end{figure}

\paragraph{Channel Permutation for Transformer Layers.} 
Applying channel permutations in Transformer architectures is challenging due to the structural coupling within multi-head self-attention (MHA) and feed-forward networks (FFN). Unlike sequential linear layers, Transformer blocks use parallel projections that share inputs and have interdependent weights, requiring coordinated, structure-aware permutations, as shown in Figure~\ref{fig:transformer_perm}.

Based on the rule that the input channel permutation of the current layer's weights will affect the output channel permutation of the previous layer's weights, the input channel permutation of \(\mathbf{W}_\mathrm{o}\) in the Transformer model will be executed as a binding to \(\mathbf{W}_\mathrm{q}\), \(\mathbf{W}_\mathrm{k}\), and \(\mathbf{W}_\mathrm{v}\). Similarly, the input channel permutation of \(\mathbf{W}_\mathrm{down}\) will be executed in a binding to \(\mathbf{W}_\mathrm{up}\) and \(\mathbf{W}_\mathrm{gate}\). 
A detailed proof can be found in supplementary materials.

\begin{figure*}[t]
\centering
\includegraphics[width=0.92\textwidth, height=0.32\textheight]{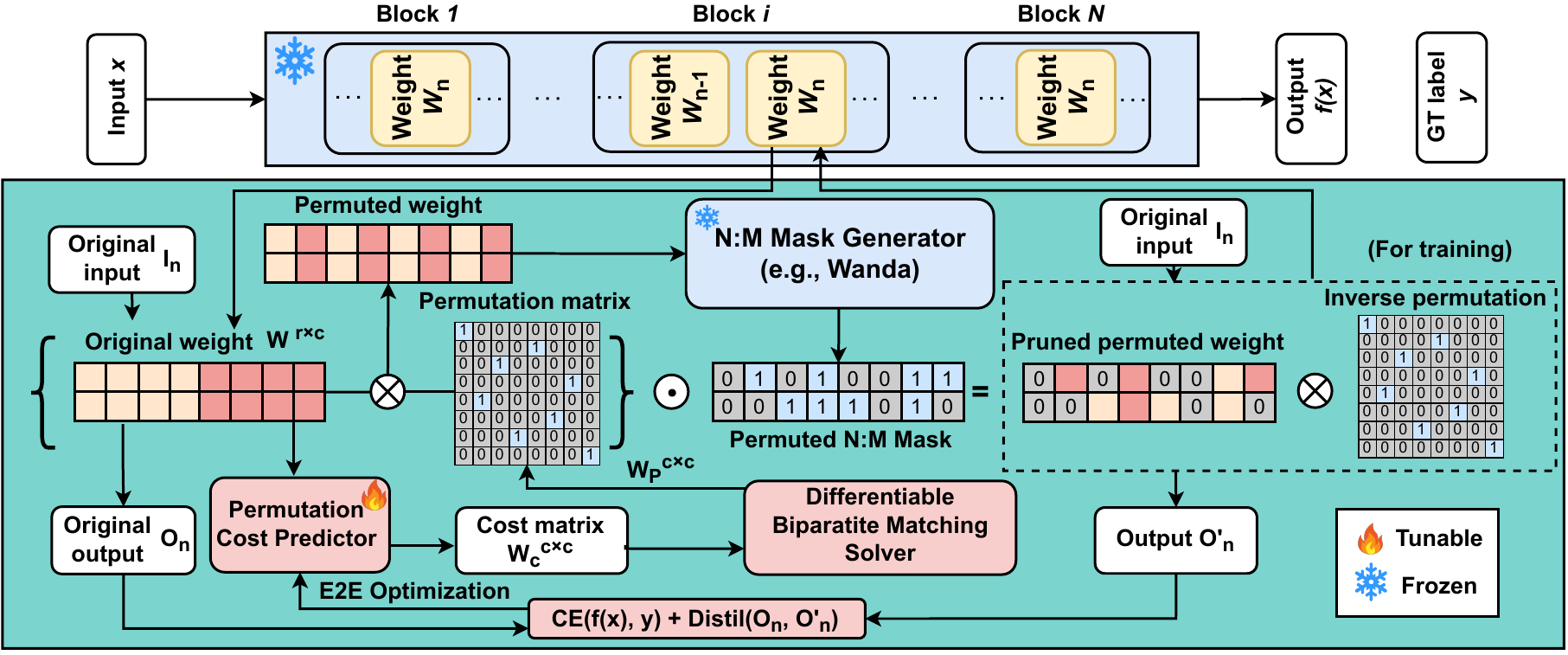} 
\caption{Overview of our learnable permutation framework. A permutation cost predictor generates cost matrices for each linear layer, which are converted into permutation matrices via a differentiable bipartite matching solver. The original weights are permuted accordingly and then sparsified using an N:M mask generator. During training, the pruned weights are inversely permuted and used for loss computation. The entire process supports end-to-end optimization while maintaining gradient flow through the binary permutation matrix generated by the differentiable solver.}
\label{fig:overview}
\end{figure*}


\subsection{Learnable Channel Permutator}
\label{sec:perm}
To support semi-structured N:M sparsity constraint in Transformer models, we propose a learnable channel permutation framework that enables end-to-end optimization of permutation operators. 
As shown in Figure~\ref{fig:overview}, the framework consists of a permutation cost predictor, a differentiable bipartite matching solver, and an optimization training objective. The cost predictor produces layer-wise cost matrices, which are used to generate permutation matrices. These reorder the weights before pruning by an N:M mask generator (actually structured sparsity pruning method, we use Wanda in this paper). During training, pruned weights are inversely permuted for loss computation, preserving gradient flow through the pipeline.
Unlike heuristic approaches based on local importance or static ranking, our method learns permutation jointly across gradient-based end-to-end optimization. 


\paragraph{Permutation Cost Predictor.}
\label{sec:cost}
The core of our method is a learnable permutation cost predictor, which produces a cost matrix to guide the reordering of channels or features. 
Given a weight matrix \(\mathbf{W} \in \mathbb{R}^{d_{\text{out}} \times d_{\text{in}}}\) from a weight matrix, the goal is to construct a permutation matrix \(\mathbf{P} \in \{0,1\}^{d_{\text{in}} \times d_{\text{in}}}\) that rearranges the input channels such that the pruned model aligns more effectively with structured N:M sparsity constraints.

To enable differentiable learning of \(\mathbf{P}\), we introduce a real-valued cost matrix \(\mathbf{C} \in \mathbb{R}^{d_{\text{in}} \times d_{\text{in}}}\), where each element \(C_{i,j}\) reflects the cost of assigning the original input channel \(i\) to position \(j\). Intuitively, \(\mathbf{C}\) encodes the pairwise preference for spatial relocation, integrating both structured sparsity alignment and semantic preservation objectives.
We parameterize the cost matrix using  learnable parameters. For each input channel, we implement a \(d_{\text{in}} \times d_{\text{in}}\) learnable parameter as the cost predictor. 
The predictor outputs a normalized cost matrix, which quantifies the cost of swapping two input channels. We minimize the cumulative cost of each cost matrix with our proposed  bipartite matching solver. Our experimental results (Table~\ref{tab:vit_results}) demonstrate that the proposed permutation cost predictor achieves strong pruning performance despite its simple design, which is intentionally lightweight to minimize additional training overhead.

Furthermore, directly learning full permutation matrices for large-dimensional weight tensors in Transformer models is computationally expensive and often unnecessary. To improve scalability and reduce overhead, we adopt a group-wise permutation strategy, where the input channels of each layer are partitioned into non-overlapping groups size \(G\), and a separate permutation is learned within each group. 

\paragraph{Differentiable Bipartite Matching Solver.}
\label{sec:match}

To enable gradient-based optimization of permutation matrices, we introduce a differentiable bipartite matching solver. Given a learned cost matrix \(\mathbf{C} \in \mathbb{R}^{N \times N}\), where \(\mathbf{C}_{i,j}\) indicates the cost of mapping the \(i\)th input channel to the \(j\)th output, our goal is to find a permutation matrix \(\mathbf{P} \in \mathcal{P}\) minimizing:
\begin{equation}
    \min_{\mathbf{P} \in \mathcal{P}} \langle \mathbf{C}, \mathbf{P} \rangle,
\end{equation}
where \(\mathcal{P}\) is the set of all \(N \times N\) permutation matrices.

Since \(\mathcal{P}\) is discrete and non-differentiable, we relax the optimization over the Birkhoff polytope \(\mathcal{B}_N\)—the convex hull of \(\mathcal{P}\)—which comprises all doubly stochastic matrices:
\begin{equation}
\mathcal{B}_N = \left\{ \mathbf{P} \in \mathbb{R}^{N \times N} \,\middle|\, \mathbf{P} \ge 0,\; \mathbf{P} \mathbf{1} = \mathbf{1},\; \mathbf{P}^{\top} \mathbf{1} = \mathbf{1} \right\}.
\end{equation}

We adopt an entropy-regularized formulation to approximate soft permutations within the Birkhoff polytope \(\mathcal{B}_N\), solved via Sinkhorn iterations~\cite{mena2018learning}:
\begin{equation}
\min_{\mathbf{P} \in \mathcal{B}_N} \langle \mathbf{C}, \mathbf{P} \rangle + \varepsilon \sum_{i,j} P_{ij}(\log P_{ij} - 1),
\end{equation}
where \(\varepsilon\) is the temperature parameter controlling the entropy strength. The solution has a closed-form structure \(\mathbf{P} = \mathrm{Diag}(u)\, \mathbf{K}\, \mathrm{Diag}(v)\), with \(\mathbf{K} = \exp(-\mathbf{C}/\varepsilon)\), and the scaling vectors \(u\) and \(v\) are iteratively updated (in the log domain) to ensure numerical stability and convergence.

This soft matching mechanism, known as SinkhornPop~\cite{knight2008sinkhorn,mena2018learning}, provides a differentiable and numerically stable approximation of the optimal permutation. It circumvents the need for non-differentiable alternatives such as the Hungarian algorithm combined with straight-through estimation (STE). During training, the temperature \(\varepsilon\) is gradually annealed to sharpen the relaxed permutation matrix towards discreteness. At inference time, the final discrete permutation is recovered by solving the original assignment problem using the Hungarian algorithm.

\paragraph{End-to-End Optimization Objectives.}

To jointly optimize channel permutation for structured pruning, we optimize the framework using a composite loss that combines task-level supervision with intermediate feature alignment. Specifically, the total objective includes two components: a \textit{task-level cross-entropy loss} and a \textit{layer-wise distillation loss}, encouraging both strong downstream performance and internal structural consistency.

The task-level loss directly optimizes the pruned model’s predictions. Let \(f_{\text{perm+prune}}(x)\) denote the output of the model after applying the learned permutation and N:M structured pruning. The cross-entropy loss is given by:
\begin{equation}
\mathcal{L}_{\text{task}} = \mathrm{CE}(f_{\text{perm+prune}}(x),\, y),
\label{eq:11}
\end{equation}
where \(\mathrm{CE}(\cdot)\) is the cross-entropy loss and $y$ is the ground-truth label. This objective ensures the learned permutation contributes to preserving task-level accuracy.

To preserve semantic consistency at the feature level, we incorporate a layer-wise distillation loss that aligns intermediate representations between the original and pruned models. Let \(h_l^{\text{orig}}\) and \(h_l^{\text{perm+prune}}\) denote the output features of the $l$-th layer in the original and permuted-pruned models, respectively. The distillation loss is defined as:
\begin{equation}
\mathcal{L}_{\text{distill}} = \sum_{l=1}^{L}  \| h_l^{\text{orig}} - h_l^{\text{perm+prune}} \|_2^2,
\label{eq:12}
\end{equation}
where $L$ is the number of pruned layers. This loss promotes the retention of task-relevant features despite structural modifications. And the final training objective combines both losses as:
\begin{equation}
    \mathcal{L}_{\text{total}} = \mathcal{L}_{\text{task}} + \alpha \, \mathcal{L}_{\text{distill}},
    \label{eq:13}
\end{equation}
where $\alpha$ balances the two losses. This joint objective provides both global supervision and local guidance for effective permutation learning under structured sparsity.

\subsection{Training and Inference Details}

\paragraph{Training.}  
We begin by collecting input feature statistics for each layer of the pretrained Transformer, which are later used by the structured pruning method (Wanda) adopted in our framework. After this preprocessing stage, all Transformer weights are frozen, and only the parameters of the permutation cost predictor remain trainable.

To preserve structural consistency in attention layers, we impose a synchronized permutation constraint across coupled projection matrices, such as \(\mathbf{W}_q\), \(\mathbf{W}_k\), and \(\mathbf{W}_v\). These matrices share the same input representation, and apply consistent permutations. 
By enforcing a shared permutation across structurally dependent components, our method ensures correctness and preserves the potential for real-world acceleration under structured sparsity.

\paragraph{Inference.}  
At inference time, the learned permutation cost predictor is frozen. For each group of channels, we can obtain the optimal permutation matrix via our bipartite matching solver. These permutations are applied to reorder the weights of each layer, followed by the application of N:M sparsity masks generated by the structured pruning method (Wanda). The resulting pruned model, with permuted and sparsified weights, is then used for standard  inference.

\section{Experiments}


\begin{table}[]
\centering
\resizebox{0.4\textwidth}{!}{
\begin{tabular}{c|c|c|cc}
\toprule
\textbf{Model} & \textbf{Sparsity} & \textbf{Method} & \textbf{Top-1 (\%)} & \textbf{Top-5 (\%)} \\ \midrule
\multirow{9}{*}{ViT-Base/16} & 0\% & Dense & 79.1 & 94.1 \\ \cmidrule{2-5} 
 & \multirow{4}{*}{2:4} & CP & 66.2 & 86.4 \\
 &  & Wanda & 65.8 & 86.4 \\
 &  & RIA & 66.6 & 86.6 \\
 &  & \cellcolor{gray!20}\textbf{Ours(Wanda)} & \cellcolor{gray!20}\textbf{67.9} & \cellcolor{gray!20}\textbf{87.9} \\ \cmidrule{2-5} 
 & \multirow{4}{*}{4:8} & CP & 66.8 & 88.2 \\
 &  & Wanda & 66.4 & 86.6 \\
 &  & RIA & 71.4 & 89.9 \\
 &  & \cellcolor{gray!20}\textbf{Ours(Wanda)} & \cellcolor{gray!20}\textbf{71.8} & \cellcolor{gray!20}\textbf{90.2} \\ \midrule
\multirow{9}{*}{ViT-Large/14} & 0\% & Dense & 85.2 & 97.9 \\ \cmidrule{2-5} 
 & \multirow{4}{*}{2:4} & CP & 79.5 & 94.7 \\
 &  & Wanda & 79.2 & 95.2 \\
 &  & RIA & 79.6 & 95.1 \\
 &  & \cellcolor{gray!20}\textbf{Ours(Wanda)} & \cellcolor{gray!20}\textbf{80.7} & \cellcolor{gray!20}\textbf{95.8} \\ \cmidrule{2-5} 
 & \multirow{4}{*}{4:8} & CP & 82.0 & 96.4 \\
 &  & Wanda & 82.0 & 96.4 \\
 &  & RIA & 82.3 & 96.4 \\
 &  & \cellcolor{gray!20}\textbf{Ours(Wanda)} & \cellcolor{gray!20}\textbf{82.7} & \cellcolor{gray!20}\textbf{96.5} \\ \bottomrule
\end{tabular}}
\caption{Performance of different approaches on ImageNet after pruning ViT-Base/16 and  ViT-Large/14 to the 2:4 and 4:8 structured pattern (50\,\% non-zero weights).}
\label{tab:vit_results}
\end{table}

\subsection{Experimental Setup}

\subsubsection{Models.} We evaluate our framework for structured sparsity on several Transformer backbones and task domains. For vision domain, we select \texttt{ViT-Base/16} and \texttt{ViT-Large/14}~\cite{dosovitskiy2020image} for experiments. 
For language domain, we employ \texttt{LLaMA-3.2-1B}~\cite{dubey2024llama} and \texttt{LLaMA-2-7B}~\cite{touvron2023llama} to represent the common small and large language models. For multimodal domain, we chose \texttt{Qwen2.5-VL-3B}~\cite{bai2025qwen2} as the object of study.

\subsubsection{Datasets.} For the vision Transformers, we use the canonical \texttt{ImageNet-1K} dataset~\cite{deng2009imagenet}, which consists of 1.28M training images and 50K validation images. All models are trained with the official train/validation split with standard input resolution of $224 \times 224$. 
For the language models, permutations are learned on the training set of \texttt{C4} dataset~\cite{raffel2020exploring} and Alpaca-en dataset~\cite{alpaca}, which comprises approximately 806MB of cleaned English text. For the vision-language models, permutations are learned on the training set of  Alpaca-en dataset~\cite{alpaca} and LLaVA-Instruct dataset~\cite{liu2023visual} dataset, which is a set of GPT-generated multimodal instruction-following data.

\begin{table*}[ht]
\centering
   \adjustbox{max width=0.9\textwidth}{
    {\small
\begin{tabular}{c|l|c|ccccccc|c}
\toprule
\textbf{Model} & \multicolumn{1}{c|}{\textbf{Method}} & \textbf{Wikitext2} & \textbf{Arc-Easy} & \textbf{Arc-Challenge} & \textbf{BoolQ} & \textbf{HellaSwag} & \textbf{OpenBookQA} & \textbf{WinoGrande} & \textbf{MMLU} & \textbf{Average} \\ \midrule
\multirow{8}{*}{\shortstack[c]{LLaMA‑3.2‑1B\\(2:4)}} & Dense & 9.06 & 65.36 & 31.40 & 63.82 & 47.73 & 26.60 & 60.69 & 31.19 & 46.68 \\ \cmidrule{2-11} 
 & Magnitude & 4808.42 & 27.95 & 19.45 & 38.50 & 26.13 & 11.80 & 51.78 & 23.80 & 28.49 \\
 & SparseGPT & \textbf{32.20} & \textbf{45.29} & 20.65 & 62.11 & \textbf{31.99} & 15.20 & 54.54 & 24.68 & \textbf{36.35} \\
 & Wanda & 75.76 & 37.16 & 18.17 & 62.05 & 28.57 & 12.00 & 50.20 & 24.45 & 33.23 \\
 & PrunerZero & 141.40 & 36.70 & 19.28 & 57.43 & 27.72 & 13.40 & 50.12 & 25.72 & 32.91 \\
 & CP & 68.17 & 38.35 & 18.14 & 62.08 & 28.56 & 12.40 & 53.31 & 23.82 & 33.81 \\
 & RIA & 72.56 & 38.12 & 20.33 & 61.34 & 27.12 & 12.80 & 52.76 & 24.31 & 33.83 \\
 & \cellcolor{gray!20}\textbf{Ours(Wanda)} & \cellcolor{gray!20}45.32 & \cellcolor{gray!20}42.26 & \cellcolor{gray!20}\textbf{21.08} & \cellcolor{gray!20}\textbf{62.11} & \cellcolor{gray!20}29.12 & \cellcolor{gray!20}\textbf{15.60} & \cellcolor{gray!20}\textbf{54.85} & \cellcolor{gray!20}\textbf{26.27} & \cellcolor{gray!20}35.90 \\ \midrule
\multirow{8}{*}{\shortstack[c]{LLaMA‑2‑7B\\(2:4)}} & Dense & 5.12 & 76.30 & 43.43 & 77.68 & 57.14 & 31.40 & 69.06 & 45.84 & 57.26 \\ \cmidrule{2-11} 
 & Magnitude & 52.00 & 61.87 & 30.20 & 59.79 & 45.42 & 21.80 & 60.93 & 26.87 & 43.84 \\
 & SparseGPT & 10.30 & 64.10 & \textbf{32.51} & \textbf{71.25} & \textbf{43.35} & \textbf{25.00} & \textbf{67.25} & \textbf{28.56} & \textbf{47.43} \\
 & Wanda & 11.38 & 62.75 & 30.38 & 67.65 & 41.18 & 23.60 & 62.59 & 27.82 & 45.14 \\
 & PrunerZero & 12.91 & 61.20 & 27.47 & 66.15 & 39.43 & 24.40 & 61.01 & 27.41 & 43.87 \\
 & CP & 10.68 & 63.32 & 30.96 & 66.92 & 41.32 & 23.80 & 63.56 & 26.51 & 45.20 \\
 & RIA & 10.52 & 63.67 & 31.82 & 67.13 & 42.03 & 23.00 & 64.13 & 27.56 & 45.62 \\
 & \cellcolor{gray!20}\textbf{Ours(Wanda)} & \cellcolor{gray!20}\textbf{10.17} & \cellcolor{gray!20}\textbf{64.23} & \cellcolor{gray!20}32.00 & \cellcolor{gray!20}68.17 & \cellcolor{gray!20}43.31 & \cellcolor{gray!20}23.60 & \cellcolor{gray!20}63.77 & \cellcolor{gray!20}28.13 & \cellcolor{gray!20}46.17 \\ \bottomrule
\end{tabular}}}
\caption{Comparisons of different pruning methods on the LLaMA3.2-1B and LLaMA2-7B language models. Performance across more sparsity patterns can be found in the supplementary materials.}
\label{tab:nlp_results}
\end{table*}


\subsubsection{Baselines.} Our method is compared with a variety of classic and state-of-the-art pruning baselines, including Magnitude~\cite{han2015learning}, Wanda \cite{sun2023wanda}, SparseGPT \cite{frantar2023sparsegpt}, PrunerZero~\cite{dong2024pruner}, CP \cite{pool2021channel}, and RIA \cite{zhangplug}.

\subsubsection{Evaluations.} 
For vision evaluations, we use the standard \texttt{ImageNet-1K} validation set with top-1/5 accuracy as the main metric. 
For language models, we measure perplexity on the \texttt{WikiText2} test set~\cite{merity2016pointer}. To evaluate compression effects across tasks, we report zero-shot accuracy on \texttt{ARC}~\cite{allenai:arc}, \texttt{BoolQ}~\cite{clark2019boolq}, \texttt{HellaSwag}~\cite{zellers2019hellaswag}, \texttt{OpenBookQA}~\cite{OpenBookQA2018}, and \texttt{WinoGrande}~\cite{sakaguchi2021winogrande}, along with 5-shot accuracy on \texttt{MMLU}~\cite{hendrycks2020measuring}.
For multimodal tasks, we report zero-shot accuracy on \texttt{MMMU}~\cite{yue2024mmmu}, \texttt{MMStar}~\cite{chen2024we}, and \texttt{TextVQA}~\cite{singh2019towards}.

\begin{table}[ht]
  \centering
  \scriptsize
  \setlength{\tabcolsep}{9pt}
  {\renewcommand{\arraystretch}{1.02}%
   \setlength{\aboverulesep}{0pt}\setlength{\belowrulesep}{1pt}%
      \adjustbox{max width=0.45\textwidth}{
    {\small
   \begin{tabular}{@{}c|l|ccc|c@{}}
    \toprule
    Sparsity & Method & MMMU & MMStar & TextVQA & Average \\
    \midrule
    0\% & Dense & 53.1 & 55.8 & 79.3 & 62.7 \\
    \cmidrule{1-6}
    \multirow{4}{*}{2:4}
      & Magnitude & 34.1 & 48.7 & 76.5 & 53.1 \\
      & Wanda     & 37.2 & 51.2 & 77.2 & 55.2 \\
      & RIA       & 37.3 & 51.4 & 77.1 & 55.3 \\
      & \cellcolor{gray!20}\textbf{Ours(Wanda)}
        & \cellcolor{gray!20}\textbf{38.1}
        & \cellcolor{gray!20}\textbf{51.9}
        & \cellcolor{gray!20}\textbf{77.8}
        & \cellcolor{gray!20}\textbf{55.9} \\
    \bottomrule
   \end{tabular}%
   }
   }
  }
  \caption{Performance of Qwen2.5-VL-3B under different pruning methods.}
  \label{tab:mmmu_pruning}
\end{table}

\subsubsection{Implementation Details.}
We use Wanda as our mask generator by default. Vision models are trained for 20 epochs on 2 AMD MI250 GPUs with AdamW (weight decay $0.01$, base learning rate $0.1$) under a cosine decay schedule and no warm-up, which takes around 4 hours.  Language models and vision-language models are trained for 20 and 10 update epochs, respectively, with AdamW ($\text{lr}=10^{-4}$). Training takes approximately 10 hours for a 1B model and 40 hours for a 7B model. The default permutation group number $G$ is 4. All runs use native AMP with FP16 precision, gradient accumulation set to one, and a small Smooth-L1 distillation term weighted $10^{-5}$. The default N:M sparsity is 2:4.

\subsection{Comparison with State-of-the-Art Methods}

\subsubsection{Results on Vision Transformers.}
On the vision side, we apply structured pruning to \texttt{ViT-Base/16} and \texttt{ViT-Large/14} under 2:4 and 4:8 sparsity, retaining 50\% of weights. Our method consistently achieves the highest top-1 and top-5 accuracy across all settings. For \texttt{ViT-Base/16}, it reaches 67.9\% / 87.9\% (top-1 / top-5) at 2:4 sparsity, outperforming the strong permutation baseline (RIA) by 1.3 points. At 4:8 sparsity, accuracy improves to 71.8\% / 90.2\%, ahead of RIA by 0.4 / 0.3 points. Similar gains are observed on \texttt{ViT-Large/14}, achieving 80.7\% / 95.8\% at 2:4 and 82.7\% / 96.5\% at 4:8 sparsity, both surpassing prior methods. These results demonstrate the accuracy-preserving strength of our approach on Vision Transformers.

\subsubsection{Results on Language Transformers.}
On the language side, we evaluate pruning performance on both \texttt{LLaMA-3.2-1B} and \texttt{LLaMA-2-7B} models across a range of context modeling and commonsense benchmarks. Our method—using Wanda pruning followed by a learned channel permutation—consistently improves downstream performance over baseline methods without requiring weight updates. On \texttt{LLaMA-3.2-1B}, our approach improves the average accuracy to 35.90\%, outperforming Wanda (33.23\%), CP (33.81\%), and RIA (33.83\%) by 2.1 to 2.7 points. Similar trends are observed on \texttt{LLaMA-2-7B}, where our method reaches 46.17\% average accuracy, compared to 45.14\% (Wanda), 45.20\% (CP), and 45.62\% (RIA). Notably, our method improves performance on challenging tasks such as \texttt{ARC-Challenge} (e.g., 3.79 points over Wanda) and maintains strong results on \texttt{BoolQ} and \texttt{WinoGrande}. While methods like SparseGPT achieve higher accuracy due to weight updates, our approach operates in the same constrained setting as Wanda, offering a fair and efficient comparison. Moreover, on \texttt{WikiText2}, our method yields better perplexity than baselines, demonstrating its ability to preserve language modeling quality. 

\begin{table}[t]
  \centering
  \scriptsize
  \setlength{\tabcolsep}{9pt}
  {\renewcommand{\arraystretch}{1.02}%
   \setlength{\aboverulesep}{0pt}\setlength{\belowrulesep}{1pt}%
      \adjustbox{max width=0.45\textwidth}{
    {\small
   \begin{tabular}{@{}c|c|ccc|c@{}}
    \toprule
    Epochs & Wikitext2 & Arc-Easy & Arc-Chall. & MMLU & Average \\
    \midrule
      0  & 11.38 & 62.75 & 30.38 & 27.82 & 40.32 \\
      1  & 10.56 & 62.88 & 30.89 & 27.61 & 40.46 \\
      2  & 10.31 & 63.13 & 31.06 & 27.96 & 40.72 \\
      5  & 10.27 & 64.10 & 31.91 & 28.05 & 41.36 \\
     10  & 10.21 & \textbf{64.23} & 31.88 & 28.12 & 41.42 \\
     \cellcolor{gray!20}20  & \cellcolor{gray!20}\textbf{10.17} & \cellcolor{gray!20}64.23 & \cellcolor{gray!20}\textbf{32.00} & \cellcolor{gray!20}\textbf{28.13} & \cellcolor{gray!20}\textbf{41.46} \\
    \bottomrule
   \end{tabular}%
   }
   }
  }
  \caption{Performance over training epochs on LLaMA‑2‑7B.}
  \label{tab:epoch_results}
\end{table}

\subsubsection{Results on Multimodal Transformers.}
As shown in Table~\ref{tab:mmmu_pruning}, our method achieves the highest performance on the \texttt{MMMU} benchmark using the \texttt{Qwen2.5-VL-3B} model. While baseline methods such as Magnitude, Wanda, and RIA yield scores of 34.1, 37.2, and 37.3 respectively, our approach outperforms with a score of 38.1. 
For Transformer-based models of various types, the learnable permutation consistently lifts performance beyond magnitude-only or rule-based permutation strategies, and does so while preserving the full latency advantage of $2{:}4$ sparsity.  




%

\begin{table*}[ht]
  \centering
  \footnotesize
  \setlength{\tabcolsep}{4pt}
  \renewcommand{\arraystretch}{1.15}
    \adjustbox{max width=0.9\textwidth}{
    {\small
  \begin{tabular}{@{}c|c|c|c c c c c c c|c@{}}
    \toprule
    Number of Groups & Trainable Parameters & Wikitext2 & Arc‑Easy & Arc‑Chall. & BoolQ & HellaSwag & OpenBookQA & WinoGrande & MMLU & Average \\
    \midrule
      1  & $1.0 \times$   & 10.11 & 64.90 & 32.00 & 68.01 & 43.56 & 23.80 & 63.93 & 28.07 & 46.33 \\
      2  & $0.68\times$    & 10.12 & 64.35 & 31.83 & 68.20 & 43.82 & 23.60 & 64.01 & 28.25 & 46.30 \\
  \rowcolor{gray!20}
      4  & $0.41\times$  & 10.17 & 64.23 & 32.00 & 68.17 & 43.31 & 23.60 & 63.77 & 28.13 & 46.17 \\
      8  & $0.23\times$   & 10.25 & 64.14 & 31.83 & 68.20 & 43.15 & 23.60 & 63.85 & 28.15 & 46.13 \\
     16  &   $0.12\times$  & 10.63 & 63.59 & 31.14 & 67.92 & 42.89 & 23.00 & 63.69 & 27.67 & 45.71 \\
    \bottomrule
  \end{tabular}
  }
  }
  \caption{Performance of our approach on LLaMA‑2‑7B with different permutation group numbers.}
  \label{tab:group_results}

\end{table*}

\begin{table*}[ht]
  \centering
  \footnotesize
  \setlength{\tabcolsep}{4pt}
  \renewcommand{\arraystretch}{1.15}
  \resizebox{\textwidth}{!}{
  \begin{tabular}{@{}l|l|c|c c c c c c c|c@{}}
    \toprule
    Model & Pruning Methods & Wikitext2(Ours/Baseline) & Arc‑Easy & Arc‑Chall. & BoolQ & HellaSwag & OpenBookQA & WinoGrande & MMLU & Average(Ours/Baseline) \\
    \midrule
    \multirow{4}{*}{LLaMA‑2‑7B}
      & Magnitude(Ours)    & 45.82 /52.00 & 62.88 & 30.89 & 61.68 & \textbf{45.47} & 22.20 & 62.12 & 27.12 & 44.63 /43.84 \\
      & Wanda(Ours)        & 10.17/11.38 & 64.23 & 32.00 & 68.17 & 43.31 & 23.60 & 63.77 & 28.13 & 46.17/45.14 \\
      & RIA(Ours)          & 10.02/10.52 & 63.89 & 32.51 & \textbf{68.84} & 42.59 & 24.00 & 64.33 & 29.02 & 46.45/45.62 \\
      &\cellcolor{gray!20}AdmmPruner(Ours)   &  \cellcolor{gray!20}\textbf{9.56}/9.68 & \cellcolor{gray!20}\textbf{69.02} & \cellcolor{gray!20}\textbf{32.68} & \cellcolor{gray!20}63.39 & \cellcolor{gray!20}45.12 & \cellcolor{gray!20}\textbf{26.00} & \cellcolor{gray!20}\textbf{65.98} & \cellcolor{gray!20}\textbf{29.62} & \cellcolor{gray!20}\textbf{47.40}/47.05 \\
    \bottomrule
  \end{tabular}}
  \caption{Performance of different pruning methods when integrated with our approach on LLaMA‑2‑7B.}
  \label{tab:more_baselines}
\end{table*}

\subsection{Training Convergence and Efficiency}
To assess convergence, we report performance from epoch 1 to 20 in Table~\ref{tab:epoch_results}. The initial performance is limited (average 33.23; \texttt{Wikitext2} perplexity 75.76), but improves markedly after one epoch, reaching 40.47 and 10.56 respectively, reflecting an 86\% reduction in perplexity and indicating fast optimization of the permutator. Subsequent gains taper off: 40.71 at epoch 2, 41.36 at epoch 5, 41.41 at epoch 10, and 41.44 at epoch 20.
Task-level improvements show a similar pattern. \texttt{Arc‑Easy} improves from 37.16 to 64.22, \texttt{Arc‑Challenge} from 18.17 to 31.96, and \texttt{MMLU} from 24.45 to 28.13, with most gains observed after the first epoch, which takes approximately 3 to 4 hours.

We also evaluate other heuristic permutation methods under the same settings, such as the LSA algorithm in RIA and the search-based approach in CP. These methods typically require 5 to 10 times longer to converge compared to our approach. These results suggest that the method converges efficiently, with nearly all achievable performance obtained within five epochs, only 0.08 below the final result at epoch 20, while maintaining low per-epoch computational cost.


\subsection{Ablation Study}

\subsubsection{Impact of Permutation Group Number.}

We investigate how the number of permutation groups $G$, defined as the granularity of partitioning the weight matrix before learning permutations, affects performance and efficiency. A smaller $G$ enables more global reordering, potentially leading to better permutation masks. A larger $G$ reduces the number of parameters but constrains the search space. As shown in Table~\ref{tab:group_results}, we evaluate $G \in \{1, 2, 4, 8, 16\}$ on \textsc{LLaMA‑2‑7B}.
The average score decreases only marginally from 46.33 at $G=1$ to 46.30, 46.17, 46.13, and 45.71 as the number of groups increases. \texttt{Wikitext2} perplexity rises slightly from 10.11 to 10.63. Meanwhile, the number of trainable parameters is significantly reduced.
 Considering the reduced number of trainable parameters, along with the preservation of \texttt{Wikitext2} perplexity and task accuracy, we adopt $G=4$ as the default setting in this paper. Notably, $G=8$ is also a viable option when a bigger accuracy drop is acceptable and computational resources are limited.


\begin{table}[t]
  \centering
  
  \resizebox{\columnwidth}{!}{%
    \begin{tabular}{@{}l|c|c|c|c|c@{}}
      \toprule
      Loss & Wikitext2 & Arc‑Easy & Arc‑Chall. &  MMLU & Average \\
      \midrule
      CE Loss                   & 46.15  & 42.01 & 20.53 &  26.12 & 29.55 \\
      Distillation Loss              & 52.58  & 40.51 & 20.12 &  25.97 & 28.87 \\
      \cellcolor{gray!20}CE+Distillation loss           & \cellcolor{gray!20}\textbf{45.32}  & \cellcolor{gray!20}\textbf{42.27} & \cellcolor{gray!20}\textbf{21.08} &  \cellcolor{gray!20}\textbf{26.27} & \cellcolor{gray!20}\textbf{29.87} \\
      \bottomrule
    \end{tabular}}
    \caption{Ablation of optimization loss on LLaMA3.2-1B.}
  \label{tab:loss}
\end{table}

\subsubsection{Performance over Different Pruning Methods.}

To assess the generality of our approach, we integrate the learnable permutator into four representative pruning baselines: Magnitude, Wanda, RIA, and AdmmPruner. As shown in Table~\ref{tab:more_baselines}, our method enhances performance across all settings. Specifically, the average score improves from 43.84 to 44.63 for Magnitude, from 45.14 to 46.17 for Wanda, and from 45.62 to 46.45 for RIA. The perplexity on \texttt{Wikitext2} also decreases in every case—for example, from 52.00 to 45.82 for Magnitude, from 11.38 to 10.17 for Wanda, and from 10.52 to 10.02 for RIA. On \texttt{MMLU}, the scores rise from 26.87, 27.82, and 27.56 to 27.12, 28.13, and 29.02, respectively.
Furthermore, the permutator is compatible with post-pruning weight updates. When applied to AdmmPruner, it achieves the highest overall average score (47.40), the best \texttt{MMLU} accuracy (29.62), and the lowest \texttt{Wikitext2} perplexity (9.56). These results demonstrate that the permutator is broadly applicable to both pruning baselines and post-pruning optimization techniques.

\subsubsection{Effect of Our Optimization Loss.}
To evaluate the contributions of our optimization components end-to-end cross entropy (CE) loss and layer wise distillation we conduct an ablation study. As shown in Table~\ref{tab:loss}, CE loss alone achieves an average score of 29.55 and perplexity of 46.15, while distillation alone yields 28.87. Their combination leads to the best results: an average score of 29.87 with top performance on Arc Easy (42.27), Arc Challenge (21.08), and MMLU (26.27), and a low perplexity of 45.32. These findings confirm the complementary benefits of both loss terms.


\section{Conclusion}
In this work, we present a novel end-to-end learnable permutation framework to enhance structured sparsity in large-scale transformer-based models. By introducing a differentiable permutation cost predictor and a bipartite matching solver, our approach learns optimal weight reorderings that align better with N:M sparsity constraints. 
Extensive experiments on vision and language backbones demonstrate that our method consistently outperforms state-of-the-art baselines, offering a powerful and generalizable strategy for model compression with minimal performance loss.

\bibliography{aaai2026}

\end{document}